\documentclass[runningheads]{llncs}
\usepackage[T1]{fontenc}
\usepackage{graphicx}
\usepackage{amsmath,amsfonts}
\usepackage{array}
\usepackage{textcomp}
\usepackage{stfloats}
\usepackage{url}
\usepackage{verbatim}
\usepackage{graphicx}
\usepackage{cite}
\usepackage{multirow}
\usepackage{subfigure}
\usepackage[table]{xcolor}
 \usepackage{amssymb}
 
\usepackage[misc]{ifsym}
 \usepackage{graphicx}
\usepackage{booktabs}
\usepackage{xcolor}
\begin{document}
\title{Temporal Graph Representation Learning with Adaptive Augmentation Contrastive}

%
%
\author{Hongjiang Chen\inst{1}
\and
Pengfei Jiao\inst{1} 
\and
Huijun Tang\inst{1}\footnotemark[1](\Letter) 
\and
Huaming Wu\inst{2}
}
\authorrunning{H. Chen et al.}
%
\institute{
School of Cyberspace, Hangzhou Dianzi University, Hangzhou, 310018, China 
\email{\{hchen,pjiao,tanghuijune\}@hdu.edu.cn}
\and 
Center for Applied Mathematics, Tianjin University, Tianjin, 300072, China\\
\email{whming@tju.edu.cn}
}

\toctitle{Temporal Graph Representation Learning with Adaptive Augmentation Contrastive}

\tocauthor{Hongjiang Chen, Pengfei Jiao, Huijun Tang, Huaming Wu }
\maketitle              
\renewcommand{\thefootnote}{\fnsymbol{footnote}} 
\footnotetext[1]{Corresponding Author}
\begin{abstract}
Temporal graph representation learning aims to generate low-dimensional dynamic node embeddings to capture temporal information as well as structural and property information. Current representation learning methods for temporal networks often focus on capturing fine-grained information, which may lead to the model capturing random noise instead of essential semantic information. While graph contrastive learning has shown promise in dealing with noise, it only applies to static graphs or snapshots and may not be suitable for handling time-dependent noise. To alleviate the above challenge, we propose a novel Temporal Graph representation learning with Adaptive augmentation Contrastive (TGAC) model. The adaptive augmentation on the temporal graph is made by combining prior knowledge with temporal information, and the contrastive objective function is constructed by defining the augmented inter-view contrast and intra-view contrast. To complement TGAC, we propose three adaptive augmentation strategies that modify topological features to reduce noise from the network. Our extensive experiments on various real networks demonstrate that the proposed model outperforms other temporal graph representation learning methods.

\end{abstract}
\section{Introduction}
Temporal networks have become increasingly popular for modeling complex real-world scenarios, e.g., citation networks, recommendation systems, and engineering systems \cite{holme2012temporal, hamilton2017representation, liben2003link, gorochowski2018organization}, where nodes represent interacting elements and temporal links denote their labeled interactions over time. These networks are inherently dynamic, with the topology and node properties evolving over time \cite{wang2021inductive}. However, the real world is often affected by time-varying noise, which can have a significant impact on the network structure and its predictions. For instance, colleagues who work together on a project may interact frequently during the project's duration, but may rarely interact afterwards, leading to a decrease in the amount of available information for future interactions. Therefore, it is imperative to investigate techniques for reducing the influence of time-varying noise on temporal graphs in order to improve the accuracy of predicting future interactions.

In recent years, there has been a surge in the development of temporal graph neural networks (TGNNs), which extend the capabilities of neural networks to structured inputs and have achieved state-of-the-art (SOTA) performance in various tasks, such as link prediction. However, one of the key challenges in temporal graph representation learning is the presence of time-varying noise, which can significantly affect the network's evolution. Existing methods~\cite{wen2022trend, liu2022neural, huang2022learning, liu2021inductive, jin2022neural} have primarily focused on capturing fine-grained information to obtain a more comprehensive node representation. This can lead to overfitting and the capture of random noise, which can obscure essential semantic information in the network as it evolves. Therefore, it is important to explore new approaches that balance the capture of both fine-grained and essential semantic information in order to improve the robustness and generalization ability of TGNNs.

Contrastive learning (CL) has emerged as a promising approach for addressing the aforementioned challenges in temporal graph representation learning by enabling the method to learn more generalized graph representations through the generation of multiple views for each instance using various data augmentations. This process helps reduce the impact of noise and improve method generalization and robustness \cite{zhu2020deep}. However, current graph augmentation methods tend to focus primarily on capturing structural features at the node or graph level, while neglecting the temporal information of edge generation \cite{wu2021self}. Incorporating temporal information related to edge generation into graph learning can help capture the dynamic evolution of the graph and improve the accuracy of node representations. Thus, there is a need to develop new approaches that effectively integrate temporal information into CL-based methods for temporal graph representation learning.

\begin{figure}[th!]
    \centering
    \vspace{-0.1in}
    \includegraphics[width=0.9\textwidth]{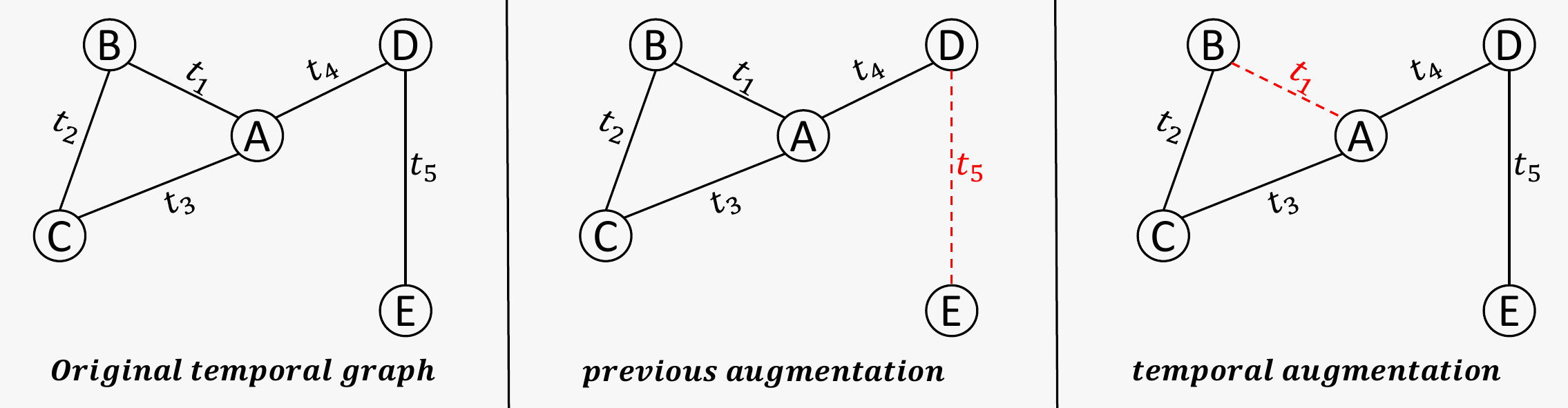}
    \caption{The toy example illustrates the limitations of the static graph augmentation method when applied to a temporal graph. Specifically, the original temporal network (left) and the resulting loss of temporal information following the application of static graph augmentation (middle) are demonstrated.
    To address this issue, we propose a novel approach for augmenting temporal graphs by incorporating both topological and temporal information. This approach allows us to eliminate redundant information while preserving vital temporal information (right).}
    \label{fig: example}
\end{figure}

Consider the toy example of a temporal network shown in Fig.~\ref{fig: example}. When using the method of static graph augmentation (e.g., GCA~\cite{zhu2021graph}) to improve the temporal graph, the edge between nodes D and E may be inadvertently removed. As a result, TGNNs may not be able to accurately predict future interactions based on the enhanced graph because crucial temporal information has been lost. Specifically, the interaction between nodes D and E at the most recent time $t_5$ is crucial for accurately predicting future interactions, while the interaction between nodes B and C at time $t_2$ may be less important. Consequently, the static graph augmentation method fails to capture important temporal information that is essential for accurate predictions of future interactions in temporal graphs. To overcome this issue, incorporating temporal information into data augmentation and node representation can effectively capture the evolution of edge generation and improve the accuracy of future interaction predictions.

In this paper, we propose a novel contrastive model called Temporal Graph representation learning with Adaptive augmentation Contrastive (TGAC). Firstly, we utilize centrality measures to eliminate redundant topological information from the input temporal graph by taking into account both structural and temporal influence. This process enhances the effectiveness of temporal graph augmentation. Subsequently, the pruned graph is subjected to perturbations to generate two distinct temporal views for augmentation. Finally, the model is trained using a contrastive loss function to maximize the agreement between node embeddings in the two views. 

Specifically, the main contributions are summarized as follows.
\begin{itemize}
    \item We present a novel approach for temporal graph contrast learning that incorporates temporal information during edge generation. This enables the model to better capture the structural evolution characteristics of graphs, resulting in improved representation learning.
    \item We propose a temporal graph augmentation method that leverages both the structural and temporal information of neighborhoods. By doing so, we are able to augment the original graph while preserving important temporal features.
    \item To further enhance important topology structures and improve node representations, we propose a graph pruning scheme that employs edge centrality measures to remove noisy or redundant connections prior to attention allocation.
    \item Experimental results demonstrate the superior performance of our proposed TGAC in tasks such as link prediction and node classification, when compared to other state-of-the-art temporal graph representation learning models. 
\end{itemize}

\section{Related Work}
In this section, we will provide a concise overview of the existing literature on temporal graph representation learning. We will then delve into the topic of contrastive representation learning methods. Finally, we will compare and contrast our proposed method with related works in the field to better understand its unique contributions.

\subsection{Temporal Graph Representation Learning}

Graph representation learning methodologies are designed to generate embeddings that capture both structural and attribute information at either the node or graph level~\cite{grover2016node2vec,wang2016structural,tang2015line,perozzi2014deepwalk}. For temporal graphs, traditional representations can be expanded to incorporate time-dependency, where the model of temporal dependence is formulated either as snapshot-based or event-based methods~\cite{longa2023graph}. These techniques aim to learn temporal node or graph embeddings that capture the evolution of the graph over time. While snapshot-based paradigms may have merit, our paper focuses primarily on event-based models, which have exhibited superior performance in empirical studies compared to models based on snapshot temporal graphs~\cite{tian2021self}.

Temporal graphs exhibit the time-varying behavior of nodes, which provides distinct insights not present in static graphs. By incorporating historical interaction information, we can distinguish between nodes that have similar local neighborhoods but different structural roles. 
For instance, JODIE \cite{kumar2019predicting} learns the embeddings of evolving trajectories by leveraging past interactions. TGN \cite{rossi2020temporal} keeps track of a memory state for each node and updates it with new interactions.
CAWs~\cite{wang2021inductive} capture the dynamic evolution of networks by using temporally anonymous random walks to extract temporal network motifs. Unfortunately, all of the aforementioned techniques do not take into account the impact of noise in the network, which can be detrimental to the ability to capture valuable temporal information.

\subsection{Contrastive Representation Learning}
Inspired by recent advancements of CL in computer vision \cite{jing2020self} and natural language processing \cite{liu2021self} domains, some research has been conducted to apply CL to graph data. For instance, DGI \cite{velickovic2019deep} combines Graph Neural Networks with infomax and concentrates on contrasting views at the node level by generating multiple augmented graphs through handcrafted augmentations. GRACE \cite{zhu2020deep} generates two views by randomly masking node attributes and removing edges, while GCA \cite{zhu2021graph} employs a similar framework to GRACE but emphasizes designing the adaptive augmentation strategy. 

Although some studies have explored the potential of contrastive learning for temporal graphs, most of them focus on static graphs and snapshot-based temporal graphs \cite{gutmann2012noise, peng2020graph, park2022cgc}. In contrast, our proposed approach addresses the challenge of noise in temporal graphs by considering the importance of edges with respect to both temporal and topological features, and adaptively augmenting the graphs in an efficient manner. Our approach effectively enhances both the temporal and topological features of the graphs, distinguishing it from existing methods for temporal graph learning and graph contrastive learning.

\section{The Proposed Method}

In this section, we will introduce the notations and definitions used in this paper. Then, we will present the problem formulation and introduce the overall framework of TGAC. Finally, we will provide a detailed description of each component module.

\begin{figure*}[tbp]
    \centering
    \includegraphics[width=0.95\textwidth]{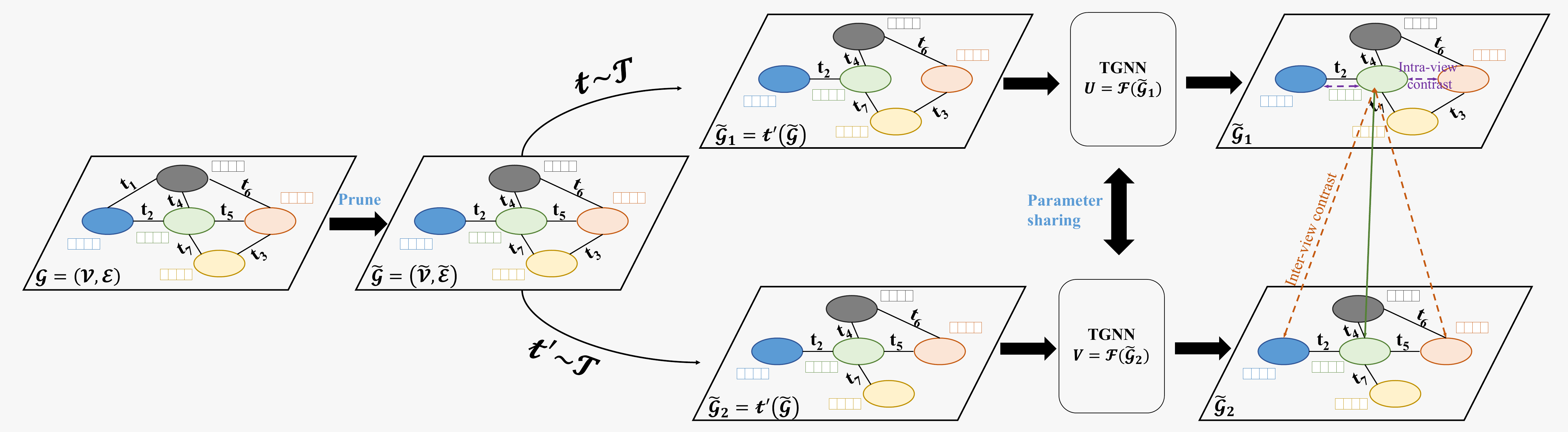}
    \caption{Our proposed Temporal Graph representation learning with Adaptive augmentation Contrastive (TGAC) model. The input graph $\mathcal{G}$ is first pruned to be $\widetilde{\mathcal{G}}$, then use two augmentation $t$ and $t'$ are generate two temporal graphs $\widetilde{\mathcal{G}}_1$ and $\widetilde{\mathcal{G}}_2$. A shared TGNN $\mathcal{F}$ is employed to obtain two views' node representation. Finally, the model was trained by contrasting positive-negative pairs in both intra-view (in purple) and inter-view (in orange).}
    \label{fig: work}
\end{figure*}

\subsection{Preliminaries}

First, we define the temporal graph based on the timestamps accompanying the node interactions.
\vspace{-0.03in}
\begin{definition}
    [Temporal Graph]
    A temporal graph is represented as $\mathcal{G} = (\mathcal{V}, \mathcal{E})$, where $\mathcal{V}$ is the set of nodes and $\mathcal{E}$ is the set of sequences of node interactions with timestamps labels. For any edge $(u, v, t)\in \mathcal{E}$, there exists a set of timestamps $\mathcal{E}_{u, v} = {(u, v, t_1), (u, v, t_2), \cdots, (u, v, t_n)}$, indicating that nodes $u$ and $v$ have interacted at least once at each of the corresponding timestamps. Two interacting nodes are referred to as neighbors. It is important to note that in temporal graphs, the concept of interaction replaces the concept of edges, and multiple interactions can occur between two nodes.
\end{definition}

\vspace{-0.03in}
 A good representation learning method for temporal networks should be able to accurately and efficiently predict how these networks will evolve over time. In this context, the problem can be formulated as follows.
\begin{definition} [Problem formulation] 
For any temporal graph $\mathcal{G} = (\mathcal{V}, \mathcal{E} ) $, the task is to learn the mapping function $f: \mathcal{V} \rightarrow  \mathbb{R}^d$ to embed the node in a d-dimensional vector space, where  $d \ll \left |\mathcal{V} \right |$.The node representation is supposed to contain both structural and temporal information and is suitable for downstream machine-learning tasks such as link prediction, and node classification.
\end{definition}

\subsection{Overview}
The proposed model utilizes graph contrastive learning to capture the structural and temporal features from temporal graphs during the training phase. The model prunes the input temporal graph, generates contrasting views, and uses a loss function that includes both link prediction and contrastive loss to learn effective node representations.

\subsection{Temporal Graph Pruning}

To ensure effective node representation learning for downstream tasks, it is necessary to remove noisy edges from the original time graph topology. TGAC achieves this by computing the importance of each edge, which takes into account both the node centrality and temporal information. As a result, the pruned time graph provides richer information for TGNN to learn node representations more effectively. The centrality of each edge is assessed based on a combination of node properties, graph topology, and temporal characteristics. By removing noisy links based on their centrality attributes, the pruned temporal graph facilitates improved information use for node representation learning through TGNN.

Node centrality is a common method for measuring the importance of nodes in large-scale complex networks. Various techniques have been proposed to measure node centrality, some of which are outlined below:

\begin{itemize}
\item Degree centrality (DE) is considered one of the elementary measures of centrality, which quantifies the number of edges incident to a particular node in a network. It is a widely used and effective approach for evaluating the significance of a node in a network. Specifically, in social networks such as Twitter, nodes represent people, while edges represent the following connections among them. Nodes with a high degree of centrality tend to correspond to more important people.

\item Eigenvector centrality (EV) is another important centrality measure that considers not only the number of connections of a particular node but also the centrality of its neighboring nodes. The idea is that if a node is connected to other nodes with high centrality, its own centrality is subsequently augmented. Consequently, a node's eigenvector centrality may not necessarily be high even if it has a substantial degree, in cases where all its connections have low centrality.
Subsequent paragraphs, however, are indented.

\item PageRank centrality (PR) is a measure of centrality determined by utilizing the PageRank algorithm. This algorithm involves developing a random walk model on a directed graph and calculating the likelihood of visiting each node under specific conditions. The resulting stable probability value of each node is its PageRank value, which serves as an indicator of the node's importance or centrality within the network.

\end{itemize}

These three methods of calculating node centrality have distinct advantages and limitations. DE is a straightforward and efficient method, making it suitable for datasets that are not very sensitive to node characteristics. EV takes into account both node characteristics and topology and performs well across a wide range of datasets. PR is especially effective for analyzing complex topological networks. Consequently, we can use the notation $\varphi(\cdot)$ to indicate the specific node centrality method used for a given dataset.

Additionally, since temporal graphs contain temporal information that static graphs lack, we need to consider the temporal dimension when measuring the impact of each edge. To achieve this, we define the centrality of each edge as $\phi^{t}_{uv}$, which is determined by the centralities of the two nodes it connects and the time of its occurrence. In undirected graphs, $\phi^{t}{uv}$ is computed as the product of the average centralities of its two nodes and the time at which the edge is formed. This can be expressed mathematically as follows:
\begin{equation}     
\phi^{t}_{uv} = (\varphi(u) +\varphi(v)) / 2 + \alpha t_{uv}.  
\end{equation}
 
This definition enables us to capture the evolving nature of the temporal graph and obtain more precise node representations that can be used for downstream tasks. In the case of a directed graph, we define the centrality of an edge as the product of the centrality of the node it is pointing to and the time at which the connection is established. This reflects the impact of the edge in directing the flow of information or influences toward the target node, while also taking into account the time factor. Hence, the edge centrality for a directed graph is defined as:
\begin{equation}   
\phi^{t}_{uv} = \varphi(v)  + \alpha t_{uv}.
\end{equation}

After obtaining the centrality score for each edge, we sort all the edges in descending order based on their centrality scores and then select the top $k$ edges to retain while pruning the rest. The value of $k$ is determined by the formula $k = E \times (1-c)$, where $E$ represents the total number of edges in the temporal graph, and $c$ is the pruning ratio. The temporal graph after pruning is illustrated below:
\begin{equation}
    \widetilde{\mathcal{E}} = \left\{u_i, v_i, t_i | \phi^{t_i}_{u_iv_i}  \in \operatorname{TopK}(\phi(\mathcal{E}), k)\right\}.
    \label{eq:prune}
\end{equation}

This method helps to remove redundant and noisy edges from the temporal graph and obtain a pruned temporal graph that can be used for subsequent training, which facilitates the acquisition of improved representation results.

\subsection{Temporal Graph Encoder}

The temporal graph encoder is based on TGN~\cite{rossi2020temporal} and consists of interchangeable and independent modules. Each node in the model has a memory vector that represents its past interactions in a compressed form. When a new event occurs, the mailbox module calculates the message for each related node, which is then used to update the node's memory vector. To address the issue of stale information, the embedding module calculates node embeddings at each time step by using their neighborhood and memory state. In other words, the encoder updates the memory state of each node with new interactions and employs a node memory update mechanism. In the node memory storage module, at time $t$, the model stores the memory of each node $u$ it has encountered so far in a vector denoted by $s_u(t)$. Whenever a new interaction occurs with a node, its compressed historical information is used to update its memory state. During the message passing and updating phase, the model calculates the memory vectors for the source and target nodes $u$ and $v$ affected by each event. This is done using the $msg$ method, which computes the message sent from the source node to the target node. The message is then used to update the memory vectors of both nodes. We formulate message passing function as
\begin{align}
    {m}_u(t) &= {msg}\left({s}_u(t^-), {s}_v(t^-), t\right), \\
    \bar{m}_u(t) &= {agg}\left({m}_u(t_1), \dots, {m}_u(t)\right). 
\end{align}

To clarify, the message passing and aggregator part involves calculating the message using the $msg$ method for the nodes $u$ and $v$ affected by each event, where $s_u(t^-)$ represents the information at node $u$ before time $t$. The message is then aggregated with the information obtained before the node, and the resulting information is then updated to yield the $s_u(t)$ value for node $u$. This process involves the utilization of a learnable information method, such as MLP, followed by information aggregation techniques, such as RNNs or attention mechanisms, and concluded with information update operations. In scenarios where nodes $u$ and $v$ are affected by an interaction event, their information is updated using a memory cell such as GRU~\cite{cho2014learning} or LSTM~\cite{hochreiter1997long}. The process can be mathematically formulated as follows:
\begin{eqnarray}
    {s}_u(t) = {mem}\left(\bar{{m}}_u(t), {s}_u(t^-)\right). 
\end{eqnarray}

Finally, after obtaining $s_u(t)$, the node representation is obtained by concatenating it with the current input features of node $u$ at time $t$, followed by a non-linear transformation to obtain the final embedding $h_u(t)$. Specifically, the concatenation operation is defined as follows:
\begin{equation} 
    {z}_u(t) = {emb}(u, t) = \sum_{v \in \mathcal{N}^k_u([0, t]) }h\left({s}_u(t), {s}_v(t)\right),
\end{equation}
where h is a learnable function. The resulting $z_u(t)$ can be used for downstream tasks such as node classification or link prediction.

\subsection{Temporal Contrastive Learning}
Contrastive learning aims to learn node or graph representations by bringing positive samples closer and pushing negative samples farther apart. We use a general contrastive learning framework to maximize representation consistency across different views. Two views of the pruned graph are generated using random augmentation operations. Existing methods struggle with topological random disturbances, as selecting positive and negative samples is crucial. After a disturbance, ineffective neighborhood information can make optimizing contrastive targets difficult. We must perturb the graph to preserve its internal mode as much as possible. Our method removes edges randomly with a probability but assigns a weight to each edge to decrease the probability of removing important edges and increase that of removing redundant ones.

To achieve this, we introduce a removal probability for each edge and improve the perturbation process of the temporal graph by considering edge importance. Similar to temporal graph pruning, we compute edge importance based on topology and time information and use it to calculate the removal probability for each edge. However, since the importance values may be relatively large, we first normalize them by setting them to $w^{t}_{uv}=\lg {\phi^{t}_{uv}}$. After normalization, we obtain the removal probability for each edge as follows:
\begin{equation}
    p^{t}_{uv} = \min\left(\frac{{w^{t}_{max}}-{w^{t}_{uv}}}{{w^{t}_{max}}-{{\mu^t_w}}}\cdot p_e, p_{r}\right),
\end{equation}
where $p_{e}$ is a hyperparameter that controls the overall probability of edge removal, $w^{t}_{max}$ and ${\mu^t_w}$ are the maximum and average values of $w^{t}_{uv}$, respectively. We set a cut-off probability $p_{r}<1$ to prevent the removal probability from becoming too high and corrupting the graph topology. The resulting temporal graph is pruned and looks like this:

\begin{equation}
    P\left\{(u,v,t) \in \widetilde{\mathcal{E}}\right\} = 1 - p^{t}_{uv}.
    \label{eq:tcl}
\end{equation}

To enhance the quality of node representations, we propose topological perturbations that generate distinct views during each iteration of training, denoted as $\widetilde{\mathcal{E}}_1$ and $\widetilde{\mathcal{E}}_2$. The probabilities of generating these two views are represented by $p^1_e$ and $p^2_e$, respectively. To prevent excessive perturbation that may lead to the degradation of the graph topology, we set $p_e$ to 0.7, ensuring that $p_r$ does not surpass 0.7.

\subsection{Loss Function}
\subsubsection{Task Loss:}
To learn the parameters of TGNN for each view node, we utilize a link prediction binary cross-entropy loss function, 
define as follows:
\begin{equation}
\begin{split}
\mathcal{L}(u, v, t) = - \log \sigma ({-{z}_{u}^{t}}^{\mathrm {T}} {{z}_{v}^{t}}) -Q\mathbb{E}_{v'\sim P(v)}\log{\sigma ({{{z}_{u}^{t}}}^\mathrm {T} {{z}_{v'}^{t}}}).
\end{split}
\end{equation}

The loss function aims to maximize the likelihood of the observed edges while minimizing the likelihood of negative edges. Since two views both have this task, the loss for the two views is defined similarly. The overall objective to be maximized is defined as the average over two views, formally given by:

\begin{align}
\label{task_loss}
\mathcal{L}_{task} = \sum_{(u_1, v_1, t_1)\in \widetilde{\mathcal{E}}_1} &\mathcal{L}(u_1, v_1, t_1) + \sum_{(u_2, v_2, t_2)\in \widetilde{\mathcal{E}}_2} \mathcal{L}(u_2, v_2, t_2).
\end{align}

\subsubsection{Contrastive Loss:}
We use a comparison objective for the two generated views to differentiate nodes with the same identifier in different views from other embeddings. For any node ${v_i}$ in one view, its corresponding node ${u_i}$ in the other view is considered as an anchor, and ${v_i}$ and ${u_i}$ form positive sample pairs. All other nodes from both views form negative samples, guiding the model to maximize the consistency of node representations across the two views. The representations of each node in the two views should be similar and distinct from those of other nodes.

Furthermore, we use a two-layer MLP to transform node representations into a feature space for comparison. A similarity function $\theta(u,v) = s(g(u), g(v))$ is used to measure different node representations, where $s$ can be either cosine or Euclidean distance and $g(\cdot)$ denotes the non-linear projection of the MLP. To achieve contrastive learning in multi-view, we use a loss function similar to InfoNCE. For each positive sample pair $u_i$ and $v_i$, the objective function is defined as follows:
\begin{equation}
\begin{split}
\label{cl_loss}
    \mathcal{L}_{cl} = \sum_{{u_i, v_i}\in \mathcal{V } } & \log \frac {P_i} {{P_i}+{N^{inter}_i} + {N^{intra}_i}},
\end{split}
\end{equation}
where $P_i = {e^{\theta\left(u_i, v_{i} \right) / \tau}}$ is positive pair, $N^{inter}_i$ and $N^{intra}_i$ are inter-view and intra-view negative pairs, respectively, which are given by the following:
\begin{align} 
    N^{inter}_i &= \sum_{k \neq i} e^{\theta\left(u_i, v_{k} \right) / \tau}, \\
    N^{intra}_i &= \sum_{k \neq i} e^{\theta\left(u_i, {u}_k \right) / \tau}, 
 \end{align}
where $\tau$ denotes the temperature coefficient.

\subsubsection{Total Loss:}

The total loss function is a combination of the task loss $\mathcal{L}_{task}$ and contrastive loss $\mathcal{L}_{cl}$. The definition of the total loss function $\mathcal{L}$ is established formally by utilizing Eqs. {\ref{task_loss}} and {\ref{cl_loss}}. Specifically, the total loss function $\mathcal{L}$ is expressed as follows: 
\begin{equation}
    \mathcal{L} = \lambda\mathcal{L}_{task} + \mathcal{L}_{cl},
    \label{total_loss}
\end{equation}
where $\lambda$ is a hyperparameter that balances the weights of the two loss functions. The task loss function $\mathcal{L}_{task}$ evaluates the predictive capability of the model in identifying observed edges in the temporal graph, whereas the contrastive loss function $\mathcal{L}_{cl}$ encourages the consistency of representations of the same node across the two augmented views. 

\section{Experiments}
In this section, we evaluate the performance of TGAC against a variety of baselines on different datasets. We further conduct an ablation study on relevant modules and hyperparameter analysis.

\vspace{-1cm}
\subsection{Experimental Setup}

\subsubsection{Datasets} 
We evaluate the performance of TGAC on the tasks of temporal link prediction and dynamic node classification using four public temporal graph datasets, namely, Wikipedia \cite{kumar2019predicting}, Reddit \cite{baumgartner2020pushshift}, MOOC \cite{kumar2019predicting}, and CollegeMsg \cite{leskovec2014snap}. 
A detailed description of the statistical characteristics of these datasets is presented in Table \ref{tab.1}.
    \vspace{-0.15in}

\begin{table}[!ht]
\setlength{\tabcolsep}{7pt}
\renewcommand{\arraystretch}{.95}
\caption{Statistics of the datasets.}
\vspace{-0.1in}
\label{tab.1}
\centering
\resizebox{0.6\textwidth}{!}{
\begin{tabular}{ccccc}
\toprule
\bfseries Datasets & \bfseries $\left |\mathcal{V} \right |$ & \bfseries $\left |\mathcal{E}  \right |$ & \bfseries Feature & \bfseries Label\\
\midrule
Wikipedia & 9,227 & 157,474  & 172 & 2\\
Reddit & 10,984 & 672,447  & 172 & 2\\
Mooc & 7,144 & 411,749 & 0 & 2 \\
CollegeMsg & 1,899 & 59,835 & 0 & 0 \\
\bottomrule
\end{tabular}}
\end{table}

\vspace{-0.5in}
\subsubsection{Baselines} To evaluate the performance of TGAC, we compare ten state-of-the-art graph embedding methods on both static and temporal graphs. 
For static graph embedding methods, including GAE, VGAE \cite{kipf2016variational}, GraphSAGE~\cite{hamilton2017inductive} and GAT~\cite{velickovic2017graph}. For temporal graph embedding methods, including CTDNE~\cite{nguyen2018continuous}, JODIE \cite{kumar2019predicting}, DyRep \cite{trivedi2019dyrep}, TGAT~\cite{xu2020inductive}, TGN~\cite{rossi2020temporal} and CAWs~\cite{wang2021inductive}.

\vspace{-0.15in}
\subsubsection{Parameter Settings} In the parameter settings, we select the optimizer with the Adam algorithm, the learning rate is 0.0001, and the dropout probability is 0.1. The dimension of both node embedding and time embedding is set to 100, memory dimension is set to 172. The temporal information weight $\alpha$ and contrastive loss weights $\lambda$ are set at 10 and 0.1. For the baseline methods, we keep their default parameter settings.

\begin{table}[!ht]
\renewcommand{\arraystretch}{.95}
\caption{ROC AUC(\%) and Average Precision(\%) for the transductive temporal link prediction on Wikipedia, Reddit, Mooc and CollegeMsg. The means and standard deviations are computed for ten runs.}
\label{tab: link}
\centering
\resizebox{\linewidth}{!}{
\begin{tabular}{llcccccccc}
    \toprule
    \multirow{2}*{Task} & \multirow{2}*{Methods} & \multicolumn{2}{c}{Wikipedia} & \multicolumn{2}{c}{Reddit} &
    \multicolumn{2}{c}{Mooc} &
    \multicolumn{2}{c}{CollegeMsg} \\
    \cline{3-10}
    ~ & ~ &  AUC & AP & AUC & AP & AUC & AP & AUC & AP \\
    \midrule
    \multirow{8}*{\rotatebox{90}{Transductive}} & GAE & $91.47\pm0.3$ & $91.12\pm0.1$ & $95.87\pm1.2$ & $96.57\pm1.0$ & $87.89\pm0.6$ & $90.70\pm0.3$ & $73.15\pm1.5$ & $70.00\pm1.17$ \\
    ~ & VGAE        & $82.43\pm1.6$ & $82.50\pm4.0$ & $92.70\pm0.4$ & $91.53\pm0.7$ & $88.21\pm0.6$ & $\textbf{91.00}\pm0.3$ & $74.07\pm0.9$ & $70.66\pm1.0$  \\
    ~ & GraphSAGE   & $92.00\pm0.3$ & $92.34\pm0.3$ & $97.75\pm0.1$ & $97.85\pm0.1$ & $56.17\pm0.3$ & $60.63\pm0.2$ & $62.38\pm1.3$ & $62.48\pm0.9$ \\
    ~ & GAT       & $92.76\pm0.5$ & $93.17\pm0.5$ & $97.90\pm0.1$ & $97.07\pm0.1$ & $67.24\pm0.1$ & $66.66\pm0.8$ & $78.09\pm0.5$ & $75.97\pm0.7$ \\
    ~ & CTDNE       & $ 82.36\pm0.7 $ & $ 80.86\pm0.7 $ & $ 85.32\pm2.0 $ & $ 87.31\pm1.4 $ & $ 88.37\pm2.6 $ & $ 89.27\pm2.0 $ & $ 81.88\pm0.7 $ & $ 80.25\pm0.8 $ \\
    ~ & JODIE       & $ 94.94\pm0.3 $ & $ 94.65\pm0.6 $ & $ 97.62\pm0.2 $ & $ 97.07\pm0.4 $ & $ 79.75\pm2.8 $ & $ 74.85\pm3.1 $ & $59.85\pm6.0$ & $54.50\pm4.4$ \\
    ~ & DyRep       & $ 94.22\pm0.2 $ & $ 94.63\pm0.2 $ & $ 98.01\pm0.1 $ & $ 98.05\pm0.1 $ & $ 80.57\pm2.1 $ & $ 77.30\pm2.2 $ & $54.75\pm6.8$ & $51.89\pm4.8$ \\
    ~ & TGAT        & $ 94.99\pm0.3 $ & $ 95.29\pm0.2 $ & $ 98.07\pm0.1 $ & $ 98.17\pm0.1 $ & $ 66.02\pm1.0 $ & $ 63.82\pm0.9 $ & $81.05\pm0.6$ & $79.16\pm0.6$ \\
    ~ & TGN         & $ 98.42\pm0.1 $ & $ 98.50\pm0.1 $ & $ 98.69\pm0.1 $ & $ 98.73\pm0.1 $ & $ \textbf{89.07}\pm1.6 $ & $ \underline{86.96\pm2.1} $ & $85.06\pm5.9$ & $85.38\pm6.4$ \\
      ~ & CAWs & $98.39\pm0.1$ & $98.62\pm0.1$	& $98.05\pm0.1$	& $98.66\pm0.1$	& $69.48\pm5.3$	& $70.11\pm6.2$	& $90.02\pm0.2$	& $92.55\pm0.1$ \\
      ~ & \cellcolor{lightgray!50} \textbf{TGAC-DE} & \cellcolor{lightgray!50} $98.85\pm0.0$	& \cellcolor{lightgray!50} $98.89\pm0.0$	& \cellcolor{lightgray!50} $98.70\pm0.0$	& \cellcolor{lightgray!50} $98.73\pm0.0$	& \cellcolor{lightgray!50} $85.39\pm1.0$	& \cellcolor{lightgray!50} $82.20\pm1.0$	& \cellcolor{lightgray!50} $91.39\pm0.6$	& \cellcolor{lightgray!50} $92.91\pm0.5$ \\
    ~ & \cellcolor{lightgray!50} \textbf{TGAC-EV} & \cellcolor{lightgray!50} $\textbf{98.86}\pm0.0$	& \cellcolor{lightgray!50} $\textbf{98.91}\pm0.0$	& \cellcolor{lightgray!50} $\underline{98.71\pm0.1}$	& \cellcolor{lightgray!50} $\underline{98.74\pm0.0}$	& \cellcolor{lightgray!50} $\underline{88.54\pm0.8}$	& \cellcolor{lightgray!50} $86.02\pm0.8$	& \cellcolor{lightgray!50} $\textbf{91.55}\pm0.7$	& \cellcolor{lightgray!50} $\textbf{93.03}\pm0.5$ \\ 
    ~ & \cellcolor{lightgray!50} \textbf{TGAC-PR} & \cellcolor{lightgray!50} $\underline{98.85\pm0.0}$	& \cellcolor{lightgray!50} $\underline{98.90\pm0.0}$	& \cellcolor{lightgray!50} $\textbf{98.76}\pm0.1$	& \cellcolor{lightgray!50} $\textbf{98.76}\pm0.1$	& \cellcolor{lightgray!50} $88.14\pm1.4$	& \cellcolor{lightgray!50} $85.47\pm1.3$	& \cellcolor{lightgray!50} $\underline{91.49\pm0.7}$	& \cellcolor{lightgray!50} $\underline{92.98\pm0.5}$ \\

    \cline{1-10}
    \multirow{8}*{\rotatebox{90}{Inductive}}    & GraphSAGE   & $88.60\pm0.3$ & $88.94\pm0.5$ & $94.28\pm0.4$ & $94.51\pm0.1$ & $53.68\pm0.4$ & $55.35\pm0.4$ & $49.64\pm1.5$ & $51.83\pm0.8$ \\
    ~ & GAT       & $89.11\pm0.5$ & $89.82\pm0.4$ & $94.30\pm0.4$ & $94.58\pm0.3$ & $53.43\pm2.1$ & $54.80\pm0.9$ & $68.98\pm1.2$ & $66.22\pm1.2$ \\
    ~ & JODIE       &
    $ 92.75\pm0.3 $ & $ 93.11\pm0.4 $ & $ 95.42\pm0.2 $ & $ 94.50\pm0.6 $ & $ 81.43\pm0.8 $ & $ 76.82\pm1.4 $ & $ 51.59\pm3.2 $ & $ 50.02\pm2.2 $ \\
    ~ & DyRep       & $ 91.03\pm0.3 $ & $ 91.96\pm0.2 $ & $ 95.79\pm0.5 $ & $ 95.75\pm0.5 $ & $ 82.06\pm1.7 $ & $ 79.17\pm1.6 $ & $ 49.05\pm4.1 $ & $ 49.30\pm2.6 $ \\
    ~ & TGAT        & $ 93.37\pm0.3 $ & $ 93.86\pm0.3 $ & $ 96.46\pm0.1 $ & $ 96.61\pm0.2 $ & $ 69.09\pm0.8 $ & $ 67.65\pm0.7 $ & $ 72.27\pm0.5 $ & $ 72.53\pm0.6 $ \\
    ~ & TGN         & $ 97.72\pm0.1 $ & $ 97.83\pm0.1 $ & $ 97.54\pm0.1 $ & $ 97.63\pm0.1 $ & $ \textbf{89.03}\pm1.6  $ & $ \textbf{86.70}\pm2.0  $ & $ 78.54\pm3.9 $ & $ 80.77\pm3.7 $ \\
      ~ & CAWs        & $98.16\pm0.2$	& $\textbf{98.52}\pm0.1$	& $97.56\pm0.1$	& $97.06\pm0.1$	& $74.79\pm2.3$	& $76.02\pm2.2$	& $\textbf{89.11}\pm1.5$	& $\textbf{91.79}\pm1.4$ \\
      ~ &\cellcolor{lightgray!50}  \textbf{TGAC-DE} &\cellcolor{lightgray!50} $\textbf{98.29}\pm0.0$	&\cellcolor{lightgray!50} $98.35\pm0.1$	&\cellcolor{lightgray!50} $\underline{98.95\pm0.0}$	&\cellcolor{lightgray!50} $\underline{98.98\pm0.0}$	&\cellcolor{lightgray!50} $84.00\pm1.3$	&\cellcolor{lightgray!50} $80.02\pm1.5$	&\cellcolor{lightgray!50} $88.42\pm0.5$	&\cellcolor{lightgray!50} $90.70\pm0.4$ \\
      ~ &\cellcolor{lightgray!50}  \textbf{TGAC-EV} &\cellcolor{lightgray!50} $\underline{98.28\pm0.1}$	&\cellcolor{lightgray!50} $\underline{98.35\pm0.1}$	&\cellcolor{lightgray!50} $98.94\pm0.1$	&\cellcolor{lightgray!50} $98.97\pm0.1$	&\cellcolor{lightgray!50} $\underline{88.23\pm0.6}$	&\cellcolor{lightgray!50} $\underline{85.30\pm0.7}$	&\cellcolor{lightgray!50} $\underline{88.49\pm0.5}$	&\cellcolor{lightgray!50} $\underline{90.75\pm0.4}$ \\
      ~ &\cellcolor{lightgray!50}  \textbf{TGAC-PR} &\cellcolor{lightgray!50} $98.28\pm0.1$	&\cellcolor{lightgray!50} $98.34\pm0.0$	&\cellcolor{lightgray!50} $\textbf{98.96}\pm0.1$	&\cellcolor{lightgray!50} $\textbf{98.98}\pm0.1$	&\cellcolor{lightgray!50} $88.16\pm1.5$	&\cellcolor{lightgray!50} $85.16\pm1.7$	&\cellcolor{lightgray!50} $88.49\pm0.5$	&\cellcolor{lightgray!50} $90.73\pm0.4$ \\ 
      
    \bottomrule
\end{tabular}}
\end{table}

\subsection{Temporal Link Prediction}

For temporal link prediction, we follow the evaluation protocols of TGN~\cite{xu2020inductive}. The goal of this task is to predict whether a temporal link will exist between given two nodes at a certain future point in time. We consider two different downstream tasks for evaluation: transductive and inductive link prediction. 
In the transductive link prediction task, we aim to predict the presence or absence of a link between two nodes that were observed during the training phase. In the inductive link prediction task, we aim to predict the presence or absence of a link between two new nodes that were not observed during the training phase. We divide the ratios of training, validation, and testing are 70\%, 15\%, and 15\%, respectively.

The results of our method and the baseline method on the temporal link prediction task are compared in Table \ref{tab: link}. We leverage the Area Under the ROC Curve (AUC) and Average Precision (AP) as performance metrics. On both transductive and inductive tasks, we make the following observations. 

\begin{itemize}
    \item   Baseline temporal graph embedding methods outperform static graph embedding methods such as GAE, VGAE, GraphSAGE, and GAT in link prediction tasks on four real-world datasets that include temporal information. 
    
    \item For the temporal graph embedding methods, compare with the methods that combine time embedding, node features, and graph topology (i.e., CTDNE, TGAT) are worse than the use of a special module to update node embeddings based on temporal interactions (i.e., TGN, TGAC). 
    
    \item Our method outperforms several existing methods on multiple datasets, although it is not as effective as CAWs on some of them. However, CAWs uses online time random walk sampling to obtain time node representations, which cannot be parallelized on the GPU and therefore require significant processing time. By incorporating prior knowledge into our time map and utilizing message passing, our method improves efficiency compared to TGN and achieves faster processing speeds than CAWs.

\end{itemize}

\subsection{Dynamic Node Classification}

For dynamic node classification, we also follow the evaluation protocols of TGN. The goal of this task is to predict the state label of the source node while giving the node link and future timestamps. Specifically, we use the model obtained from the previous transductive link prediction as the pre-training model for node classification. The node classification task trains a classifier decoder separately, such as a three-layer MLP. We evaluate the task on three datasets with dynamic node labels (i.e., Wikipedia, Reddit, and Mooc), excluding the CollegeMsg dataset because there are no node labels.

The results of our method and the baseline method on the Dynamic Node Classification task are compared in Table~\ref{tab: node}. We leverage the Area Under the ROC Curve (AUC) as performance metrics. Our results demonstrate superior performance on all three datasets, underscoring the effectiveness of our model's use of contrastive learning. By bringing the distance between nodes in one view closer while pushing away nodes in the other view, our model learns more optimized node representations for downstream classification tasks. This approach has proven to be more effective than alternative methods, as evidenced by the superior performance of our model.
 \vspace{-0.4cm}
\begin{table}[!ht]
\setlength{\tabcolsep}{7pt}
\renewcommand{\arraystretch}{.85}
\caption{ROC AUC(\%) for the transductive dynamic node classification on Wikipedia, Reddit and Mooc. The means and standard deviations are computed for ten runs. We use bold and underline to highlight the best and second best performers.} 
\label{tab: node}
\centering
    \begin{tabular}{lccc}
    \toprule
        & Wikipedia & Reddit & Mooc \\
    \midrule
    CTDNE        & $84.86\pm1.5$  & $54.38\pm7.5$ & $71.84\pm1.0$ \\
    JODIE        & $84.40\pm0.9$ & $61.51\pm1.2$ & $70.03\pm0.5$ \\
    DyRep        & $83.25\pm0.5$ & $60.86\pm1.7$ & $64.64\pm1.4$ \\
    TGAT         & $84.41\pm1.5$ & $65.98\pm1.6$ & $65.79\pm0.5$ \\
    TGN          & $87.56\pm0.7$ & $65.51\pm0.8$ & $63.93\pm0.3$ \\
    CAWs         & $84.88\pm1.3$ & $66.52\pm2.2$ & $68.77\pm0.4$\\
    \rowcolor{lightgray!50} \textbf{TGAC-DE} & $87.69\pm0.2$	& $68.54\pm0.4$ & $\underline{70.13\pm0.2}$ \\
    \rowcolor{lightgray!50} \textbf{TGAC-EV} & $\textbf{90.13}\pm0.2$	& $\textbf{71.70}\pm0.4$ & $61.83\pm0.7$ \\
    \rowcolor{lightgray!50} \textbf{TGAC-PR} & $\underline{88.85\pm0.2}$	& $\underline{71.06}\pm0.8$ & $\textbf{71.10}\pm0.3$ \\
    \bottomrule
    \end{tabular}
\end{table}
\vspace{-1cm}
\subsection{Ablation Experiment}
We conducted a series of experiments on the CollegeMsg dataset to evaluate the effectiveness of pruning on temporal graphs, using different centrality measures. Our findings, presented in Table \ref{tab: ablation}, indicate a notable enhancement in the model's performance upon the removal of extraneous links through the application of diverse node centrality principles. Herein, ``T'' refers to the TGNN function, while ``P'' denotes the Prune function. Furthermore, we conducted an ablation study to assess the impact of contrastive learning, and the results are depicted in Table \ref{tab: ablation auc}. Upon removing both the pruning and contrastive learning aspects, the model became a conventional TGN model. Our findings demonstrate that the absence of pruning and contrastive learning resulted in a significant decline in the performance of the TGN model.

\vspace{-0.4cm}
\begin{table}[!ht]
    \setlength{\tabcolsep}{7pt}
    \renewcommand{\arraystretch}{.95}
    \caption{Ablation study result on CollegeMsg for Pruning schemes}
    \label{tab: ablation}
    \centering
    \begin{tabular}{cccccccccccccc}
    \toprule
    & {T} & {T+DE} & {T+EV} & {T+PR} & {T+DE+P} & {T+EV+P} & {T+PR+P} \\
    \midrule
    AUC & 85.06 & 90.38 &  90.57 &  90.58 & \textbf{92.39} & \textbf{92.55} & \textbf{92.49}\\
    \bottomrule
    \end{tabular}
\end{table}
\begin{table*}[htp]
\renewcommand{\arraystretch}{.95}
\caption{ROC AUC(\%) for both the transductive and inductive temporal link prediction on Wikipedia, Reddit, and CollegeMsg. }
\label{tab.4}
\centering
\begin{tabular}{lcccccc}
    \toprule
    \multirow{2}*{} & \multicolumn{2}{c}{Wikipedia} & \multicolumn{2}{c}{Reddit} &
    \multicolumn{2}{c}{CollegeMsg}\\
    \cline{2-7}
    ~ &  Transductive & Inductive & Transductive & Inductive & Transductive & Inductive \\
    \midrule
    TGAC w/o CL & $98.29$ 	& $98.37$	& $98.54$	& $98.57$	& $85.06$	& $87.41$ \\
    TGAC w/o Prune & $98.32$	& $98.40$	& $98.62$	& $98.65$	& $90.57$	& $87.93$ \\
    TGAC & $\textbf{98.53}$	& $\textbf{98.64}$	& $\textbf{98.82}$	& $\textbf{98.86}$	& $\textbf{92.71}$	& $\textbf{88.79}$ \\
    \bottomrule
\end{tabular}
\label{tab: ablation auc}
\end{table*}
\begin{figure}[!ht]
\vspace{-0.5cm}\centering
\subfigure[Pruning ratio $ c$]{
    \begin{minipage}[t]{0.33\linewidth}
    \includegraphics[width=1.1\linewidth]{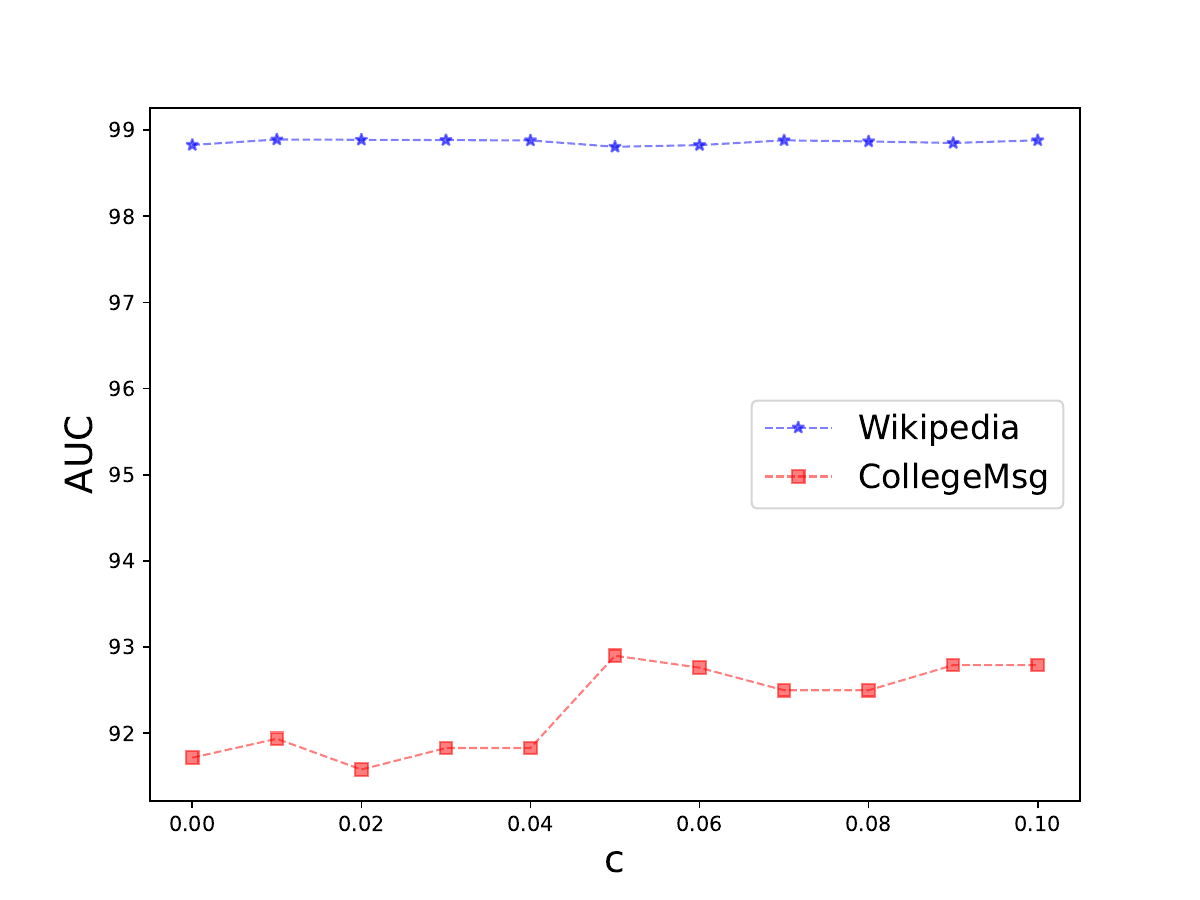}
    \end{minipage}%
}%
\subfigure[Balance weight $ \lambda$]{
    \begin{minipage}[t]{0.33\linewidth}
    \includegraphics[width=1.1\linewidth]{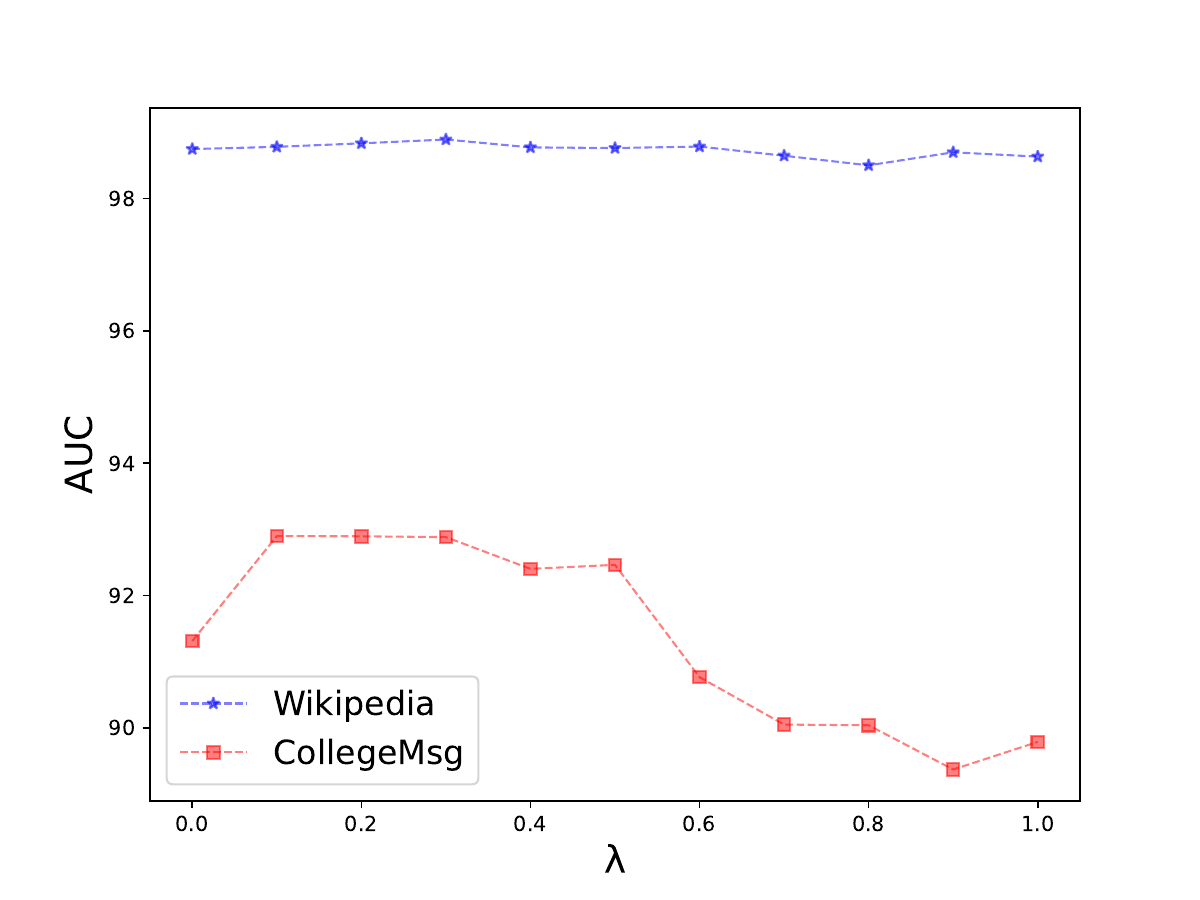}
    \end{minipage}%
}%
\subfigure[ Drop ratio $p_e$]{
    \begin{minipage}[t]{0.33\linewidth}
    \includegraphics[width=1.1\linewidth]{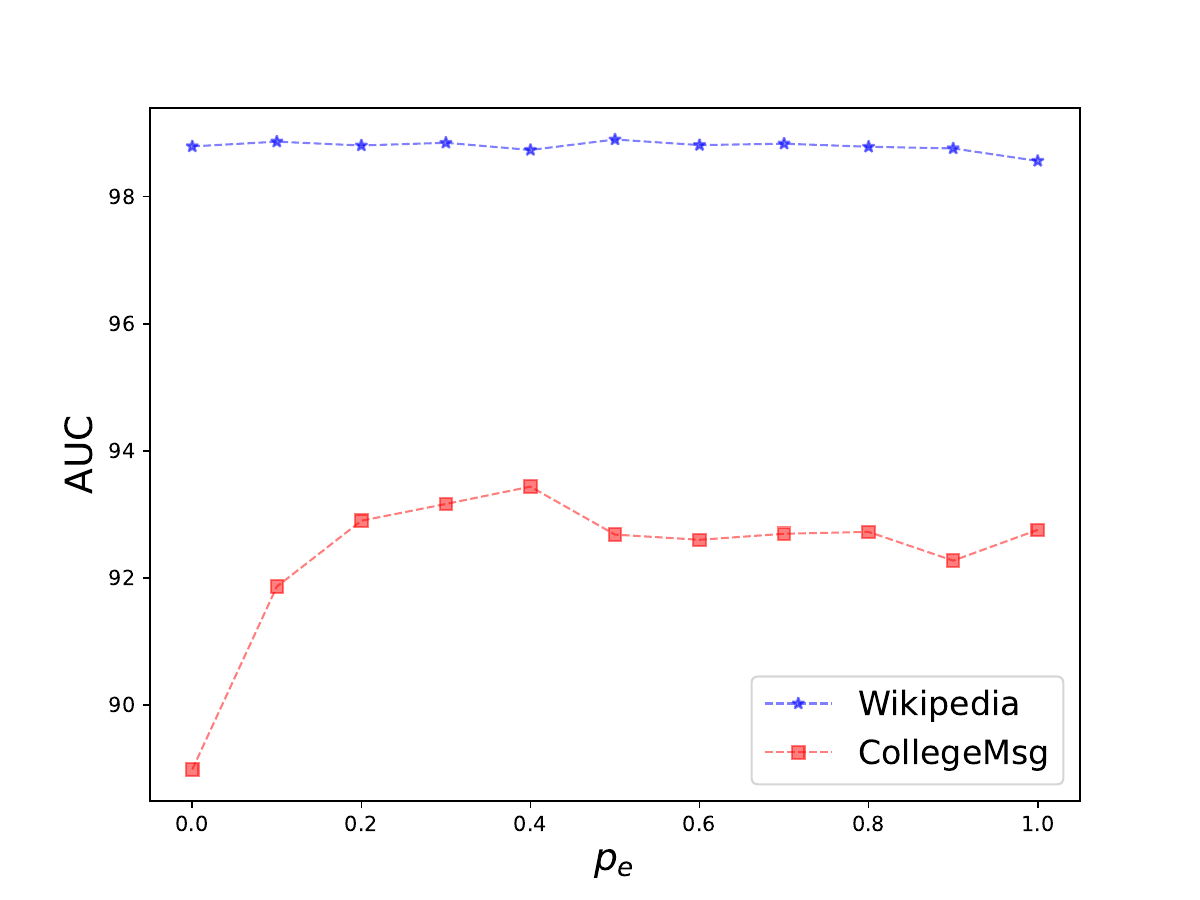}
    \end{minipage}
}%
\centering
\vspace{-0.1in}
\caption{Parameter Sensitivity.}
\label{fig:sensitivity}
\end{figure}
\vspace{-1cm}
\subsection{Parameter Sensitivity}
Our proposed method requires a thorough analysis of hyperparameters' impact on temporal link prediction performance on the datasets. These hyperparameters are the temporal graph pruning ratio $c$, the balance parameter $\lambda$, and the temporal graph enhancement factor $p_e$. We use a range of evaluation metrics to gauge the efficacy of various parameter values. We evaluate them on Wikipedia and CollegeMsg datasets using link prediction as the downstream task. We investigate the impact of the temporal graph pruning ratio on the model's ability to learn effective information. Additionally, we explore the balance between link prediction and contrastive learning. Fig.~\ref{fig:sensitivity} illustrates the sensitivity of our model's performance to various hyperparameters, including $c$, $\lambda$, and $p_e$. Our experiments show that the proposed method achieves the best results when $c=0.05$, $\lambda=0.1$, and $p_e=0.4$.

\section{Conclusion}

This paper introduces a novel temporal graph contrastive learning model named TGAC. The proposed model employs a pruning and adaptive augmentation technique that incorporates topological and temporal information with prior knowledge. This approach leads to the generation of enhanced temporal graph information, which in turn improves the performance of TGNN. The experimental results demonstrate that the TGAC model outperforms state-of-the-art methods on most of the datasets.

\section*{Acknowledgments}

This work was supported by the Fundamental Research Funds for the Provincial Universities of Zhejiang Grant GK229909299001-008 and GK239909299001-028, Zhejiang Laboratory Open Research Project under Grant K2022QA0AB01, National Natural Science Foundation of China under Grant 62071327.






\newpage

\bibliographystyle{splncs04}
\bibliography{reference}

\end{document}